\documentclass[journal]{IEEEtran}

\usepackage{graphicx} 
\usepackage{amsmath} 
\usepackage{amssymb}  
\usepackage{xcolor}
\usepackage[utf8]{inputenc}
\usepackage[english]{babel}
\usepackage[normalem]{ulem}
\usepackage{asymptote}
\usepackage[ruled]{algorithm2e}
\usepackage{algorithmic}

\ifCLASSINFOpdf
\else
\fi

\begin{document}
\title{Spherical formulation of geometric motion segmentation constraints in 
fisheye cameras}

\author{Letizia~Mariotti
        and~Ciar\'{a}n~Eising,~\IEEEmembership{Member,~IEEE}
\thanks{L. Mariotti is with Valeo Vision Systems, Tuam, County Galway, Ireland. E-mail: letizia.mariotti@valeo.com}%
\thanks{C. Eising is with the Department of Electronic and Computer Engineering, University of Limerick, Ireland. E-mail: ciaran.eising@ul.ie}
}

\markboth{
}%
{
}

\maketitle

\begin{abstract}
We introduce a visual motion segmentation method employing spherical geometry for fisheye cameras and automoated driving. Three commonly used geometric constraints in pin-hole imagery (the positive height, positive depth and epipolar constraints) are reformulated to spherical coordinates, making them invariant to specific camera configurations as long as the camera calibration is known. A fourth constraint, known as the anti-parallel constraint, is added to resolve motion-parallax ambiguity, to support the detection of moving objects undergoing parallel or near-parallel motion with respect to the host vehicle. A final constraint constraint is described, known as the spherical three-view constraint, is described though not employed in our proposed algorithm. Results are presented and analyzed that demonstrate that the proposal is an effective motion segmentation approach for direct employment on fisheye imagery.
\end{abstract}


\IEEEpeerreviewmaketitle

\section{Introduction}
\IEEEPARstart{I}{n} the automotive industry, fisheye cameras are a commonly available sensor type \cite{hughes2009}, particularly for rear view and surround view systems for human visual consumption. For Advanced Driver Assistance Systems (ADAS) and automated driving, commercial systems typically make use of forward facing, narrow field-of-view cameras. However, full $360^\circ$ scene interpretation is being increasingly investigated in more complex and short-range application spaces \cite{heimberger2017}. Of significant importance for ADAS and automated driving in general, is the detection of moving objects in the vicinity of the vehicle. 
The detection, knowledge of the location and potentially of the trajectory of a moving obstacle is essential for safe navigation. While the problem of moving object detection is almost trivial for a static observer, for a moving observer it is a significant challenge due to the apparent motion of static world features when the camera itself undergoes movement.

Fisheye lenses exhibit an extremely wide field-of-view, sometimes over $180^\circ$, which has proven to be especially useful in low speed driving applications \cite{heimberger2017}. However, the non-linear distortion introduced by the lens type cause motion to be imaged with complex patterns that are not easy to resolve. In order to solve the motion detection problem for fisheye cameras, we reformulate the problem in spherical coordinates, which is used to address both the non-linearity and the large field of view. The use of spherical coordinates simply requires a valid mapping from the fisheye image space, which is readily available if the intrinsic camera calibration parameters are known since the spherical coordinate, represented as a vector in $\mathbb{R}^3$, is the unit vector that is equivalent to a given point in image coordinates.

\begin{figure}[t]
  \centering
  \includegraphics[width=\columnwidth]{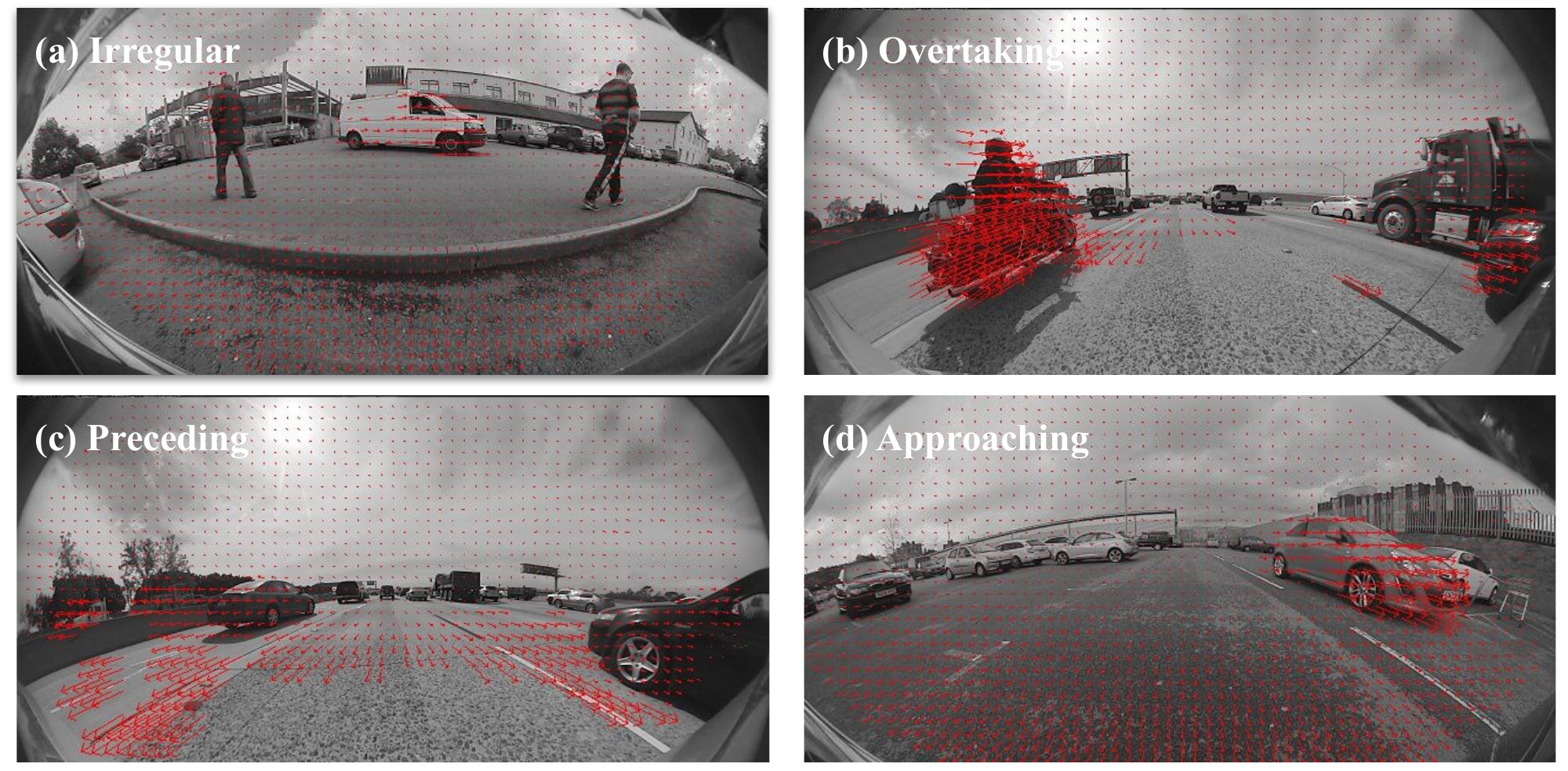}
  \vspace{-0.5cm}
  \caption{Moving objects with distinct motion feature geometry. (a) shows crossing pedestrians with irregular feature motion off the epipolar plane; (b) shows an object moving faster than the host vehicle with motion parallel to the ground plane; (c) is a vehicle moving in the same direction as the host, but with lower velocity; and (d) is an approaching vehicle with motion features parallel to the ground plane. Optical flow is overlaid in each of the images.}
  \vspace{-0.35cm}
  \label{fig::little}
\end{figure}

To solve the problem of motion segmentation using fisheye cameras, four geometric constraints (epipolar, positive depth, positive height and anti-parallel) are unified for the detection of moving obstacles in the scene.
Figure \ref{fig::little} shows the classes of motion that each of the constraints detect. Pedestrians (Figure \ref{fig::little}(a)) demonstrate irregular motion that is typically described by feature motion off the epipolar plane, so the epipolar constraint is important. However, as shall be explained, the epipolar constraint cannot differentiate motion on the epipolar plane from static features. Thus, the positive depth and positive height constraints are introduced to handle overtaking and preceding objects (Figure \ref{fig::little}(b-c)) that move largely parallel to the ground plane. There is still the class of approaching object (Figure \ref{fig::little}(d)) that cannot be detected by any of the previous constraints. The addition of the anti-parallel approach provides a more complete geometric approach to feature based moving object detection, solving this final class of motion, but at the cost of systematic false positives. For completeness, we also provide a description of the fisheye three-view constraint, though this is not implemented in our results due to the need for feature correspondences over three frames.

We structure the paper in the following way. In Section \ref{sec:background}, we augment the background material from \cite{mariotti2019} with some additional material. In Section \ref{sec::Fisheye}, fisheye mapping is described so that the reader can understand the required mapping from fisheye image space to spherical coordinates. Each of the geometric constraints is described in detail in Section \ref{sec:proposal}, with results discussed in Section \ref{sec:results}.

This work was initially presented in \cite{mariotti2019}, with significant additional material added to this paper, including a much deeper description of the geometric constraints (and describing the additional three-view constraint), adding background material in the odometry and optical flow, and providing additional description of the results, in particular a comparison with FisheyeMODNet \cite{yahiaoui2019}. Additional results are presented, for example the detection ranges in Figure \ref{fig:detRanges}. A deeper discussion is provided into how the ground truth data is generated from lidar point clouds and how the detection rates are generated.

\section{Background}
\label{sec:background}

\subsection{Related Work and Discussion}

Many methods developed for the extraction of dynamic obstacles in images have been designed to perform in specific applications, based on the appearance of the optical flow rather than a geometric understanding of the scene \cite{hultqvist2014, hariyono2014, pinto2017}. 
In contrast, in \cite{bugeau2018}, the apparent motion in the image of the background is modelled using an affine approximation, and outliers to this model are considered to be associated with a foreground moving object. This doesn't apply, however, to fisheye due to the complex distrotion of the imaged scene caused by the fisheye optics which would break the affine image motion model. In general, the limited treatment of the associated geometry in the cited studies prevents their direct application to fisheye cameras.

In the geometrical treatment of moving object detection, the use of spherical coordinates has been investigated before, especially with different formulations of the reasonably well known epipolar constraint \cite{clarke1996, soumya2012}. For a more comprehensive review of these methods, we refer to \cite{mariotti2019}. In particular, Markovi\'{c} et al. \cite{markovic2014} propose the closest to one of our approaches, describing closely the spherical epipolar constraint (\S \ref{sec:SphericalEpipolar}), though this is not explicitly mentioned. This is discussed further in the section on the spherical epipolar constraint. In addition to the epipolar constraint, Klappstein  et al. \cite{klappstein2006, klappstein2007} significantly introduced the positive depth and the positive height constraint for motion segmentation for standard field of view cameras. However, they do not translate directly to fisheye images.


Further discussion on related work is provided in \cite{mariotti2019}.

\subsection{Fisheye Mapping}

\label{sec::Fisheye}

In fisheye cameras, the lens sees with fields of view $>\!\!180^\circ$, and observed rays cannot all pass through a single flat image plane. Therefore, we cannot consider points on a projective plane, as it cannot encompass the entire FOV of a fisheye camera. Thus we will consider the representation of points on the unit sphere $\mathbf{S}^2 = \left\{\textbf{p}\in\mathbb{R}^3 \hspace{0.2cm} | \hspace{0.2cm} \|\textbf{p}\|=1\right\}$.

\begin{figure}
\begin{center}
\includegraphics[width=0.8\columnwidth]{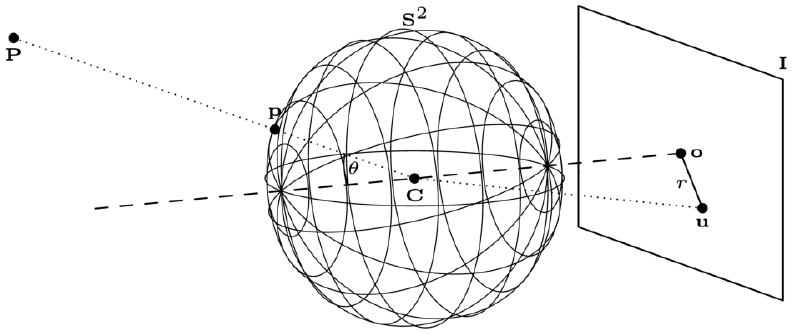}
\end{center}
\vspace{-0.3cm}
\caption{Relationship between fisheye image point and point on the unit sphere. $\mathbf{u}$ and $\mathbf{p}$ are equivalent points on the fisheye image and unit sphere respectively, with $\mathbf{p}$ laying on the same ray as $\mathbf{P}$}
\label{fig:fisheye}
\vspace{-0.5cm}
\end{figure}

What is initially required is an injective map from the image domain $\textbf{I}$ to the unit central projective sphere embedded in $\mathbb{R}^3$
$$g: \mathbf{I} \rightarrow \mathbf{S}^2$$ 
where $\mathbf{I} \subset \mathbb{R}^2$. Figure \ref{fig:fisheye} demonstrates the relationship between the image points and the unit sphere.

In principle, any appropriate definition for the mapping function $\mathbf{p} = g(\mathbf{u}), \mathbf{p} \in \mathbf{S}^2, \mathbf{u} \in \mathbf{I}$ can be used. In our case, we use a fourth order polynomial to describe the mapping of the incident angle $\theta$ to the image plane radius $r$, i.e.
\begin{equation}
    \nonumber r(\theta) = a_1\theta + a_2\theta^2 + a_3\theta^3 + a_4\theta^4
\end{equation} 
For mapping the pixel coordinate to the unit sphere, the inverse of $r(\theta)$ will be required, obtained using numerical methods.

\section{Proposed Method}

\label{sec:proposal}
In this section we describe the previously mentioned constraints in detail, with the adaptation for fisheye cameras, and specifically for spherical image coordinates.
The inputs required are, 1) displacement vectors of image points $\mathbf{u}$ and $\mathbf{u}^\prime$ between two images at two time steps (e.g. through image correspondences), and 2) the relative position of the camera and its rotation at the two time steps (e.g. through visual odometry or kinematics available on vehicle system bus). 
A summary of the described constraints is presented in Table 1 of \cite{mariotti2019}.

\subsection{Optical Flow and Odometry}

Dense optical flow is performed to give the correspondence pairs $\mathbf{u} \leftrightarrow \mathbf{u}^\prime$ in the fisheye image space $\mathbf{I}$. Optical flow continues to be an important area of research in computer vision, with recent advances in both classical \cite{kroeger2016}
and neural network methods \cite{revaud2016}. However, for computational reasons, in this paper we have elected to use the Farneb{\"a}ck algorithm \cite{farneback2003}. Of course, the methods could also be applied to sparse, feature based image registration, for example, the classical
Lucas-Kanade optical flow \cite{lucas1981} method.
To reduce the computational cost of calculations, the dense optical flow is averaged on a $5 \times 5$ pixel grid. This seems a good compromise between accuracy and noise suppression, but can be tweaked depending on the application. 

The geometry described in the next section is independent of the image registration algorithm, with the exception of the three-view constraint, which, as will be discussed, would require correspondence across three frames of video. After the image registration, the subsequent geometric processing steps are completed on the unit sphere, so the pairs $\mathbf{u} \leftrightarrow \mathbf{u}^\prime$ are raised to the points on the unit sphere $\mathbf{p} \leftrightarrow \mathbf{p}^\prime$ through the fisheye mapping discussed previously.

Odometry of the vehicle can be obtained in several ways. In computer vision, visual odometry can be directly employed, e.g. through the well known 
5-point algorithm \cite{nister2004} (in the vanilla form, requiring rectification of the image features) or other omnidirectional SLAM based methods. Alternatively, external sensors can be utilised, such as DGPS/IMU combinations, which can give highly accurate odometry estimates. However, these are not universally available on vehicles. For the results presented in this paper, we use the vehicle wheel sensors and yaw rate sensors to estimate the vehicle odometry. These are almost universally available on any modern vehicle, and can be used to obtain a full scale motion of the vehicle over the ground surface with three degrees of freedom.

Despite the fact that we use odometry without scale issues (albeit with its own problem of not quantifying non-planar motion of the vehicle), we will mention which of the geometric approaches necessarily require full scale odometry. This is important, as while visual odometry and SLAM can give very high accuracy, the issue of scale resolution is still a difficult and unsolved topic in the general case for monocular or non-overlapping camera networks \cite{liu2018}. In particular, the fisheye epipolar and the positive depth constraints do not require full scale odometry estimate. We will describe this in each case.

\subsection{Fisheye Epipolar Constraint}

\label{sec:SphericalEpipolar}

Probably the best known constraint that static points between multiple views have to satisfy is the epipolar constraint in pinhole cameras. 
Start with a point correspondence $\mathbf{p} \leftrightarrow \mathbf{p}^\prime$. In this case, the correspondence is on the projective plane (in the rest of the paper, this correspondence pair refers to points on the unit sphere). The epipolar line in the second image is given by $\mathbf{l}^\prime=\mathbf{E}\mathbf{p}, \mathbf{p}\in\mathbb{P}^2$, where $\mathbf{E}$ is the well known essential matrix (assuming calibrated camera). If the point $\mathbf{P}$ moves in space to $\mathbf{P}^\prime$, such that the image $\mathbf{p}^\prime$ of $\mathbf{P}^\prime$ does not lie on the epipolar line $\mathbf{l}^\prime$, then the constraint imposed by the essential matrix is broken, as $(\mathbf{p}^\prime)^\top \mathbf{E} \mathbf{p} \neq 0$. Geometrically, there will be a non-zero distance $d$ between $\mathbf{p}^\prime$ and the line $\mathbf{l}^\prime$ on the second image plane.

\begin{figure}[bt]
  \centering
  \includegraphics[trim={6cm 6cm 6.9cm 7cm},clip,width=\columnwidth]{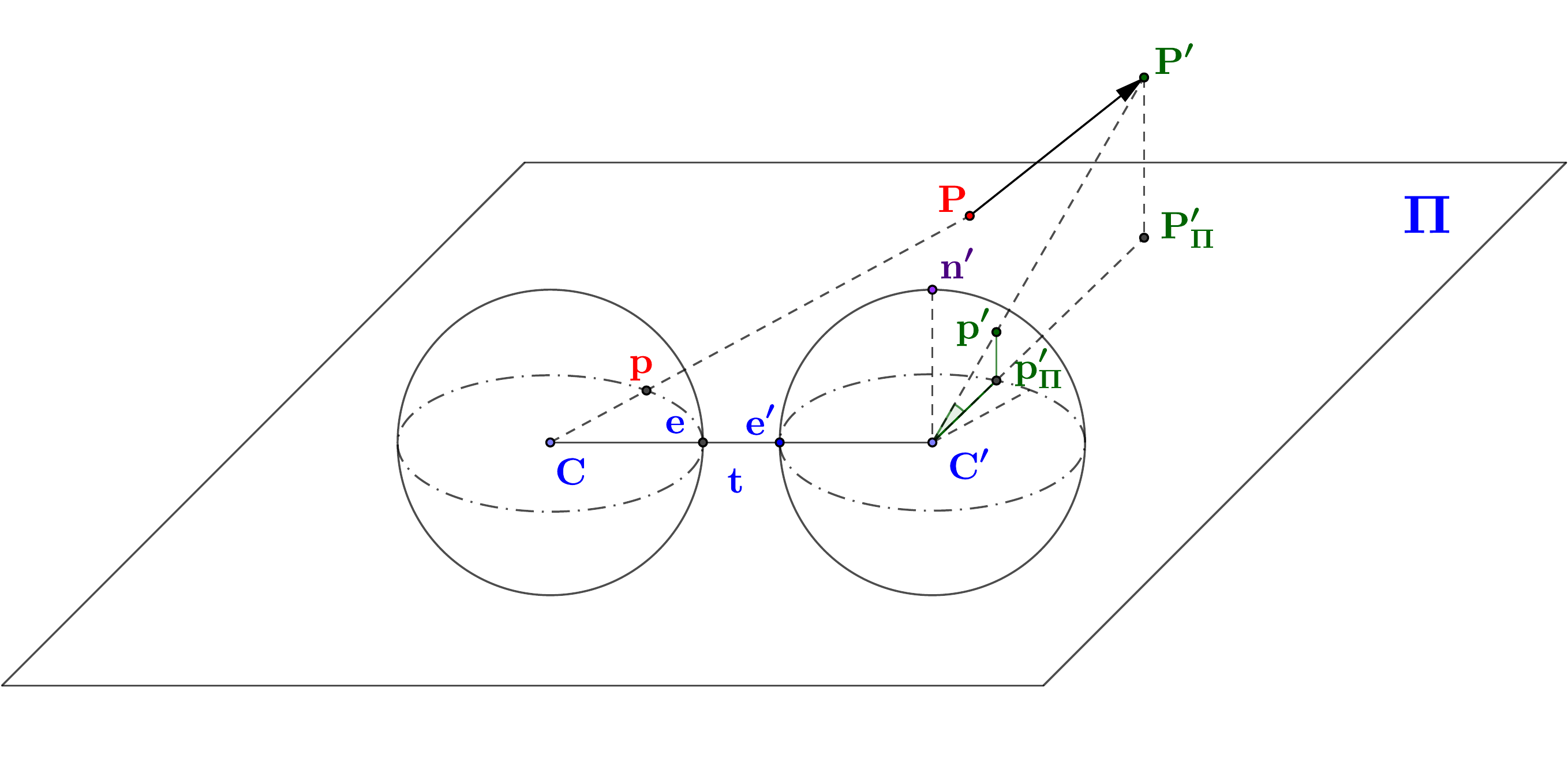}
  \vspace{-1.0cm}
  \caption{The fisheye epipolar constraint. The camera moved from $\mathbf{C}$ to $\mathbf{C}^\prime$ and the observed feature point moved from $\mathbf{P}$ to $\mathbf{P}^\prime$. The epipolar plane $\Pi$ is defined by the three points $\mathbf{C}$, $\mathbf{C}^\prime$ and $\mathbf{P}$. The vectors $\mathbf{p}$ and $\mathbf{p}^\prime$ are the projections of $\mathbf{P}$ and $\mathbf{P}^\prime$ on the unit spheres at the respective time steps. If the angle measured between $\mathbf{P}^\prime$ and its projection on the epipolar plane, $\mathbf{P}^\prime_\Pi$, is greater than zero, then the epipolar constraint is not satisfied and the absolute value of the angle is a measure of its deviation.}
  \label{fig:PlanarEpipolar}
  \vspace{-0.4cm}
\end{figure}

In fisheye imagery, however, the epipolar line is a complex curve that can be difficult to parameterise, depending on the fisheye model used. Therefore, we reformulate the restriction as the {\em fisheye epipolar constraint} and consider whether features on the unit projection sphere lie on the epipolar plane, as demonstrated in Figure \ref{fig:PlanarEpipolar}.

Fundamentally, the epipolar constraint says that viewing rays of static 3D points must meet. A 3D point that induces viewing rays that deviate from the meeting rays is violating the constraint. This constraint allows the detection of objects with a component of motion that is perpendicular to the plane of the vehicle ego-motion. This is quite similar to the method proposed by Markovi\'{c} et al. \cite{markovic2014}. While they don't explicitly define an epipolar plane, they essentially construct an epipolar great circle on the second unit sphere, and use the distance of the point in the second unit sphere to the great circle as a measure of likelihood of object motion. While achieving the same goals, we feel that our formulation is simpler.

We calculate two vectors that describe the epipolar plane for each feature point at the second time measurement. First we calculate $\mathbf{e}^\prime$, the projection on the sphere of the translation vector between the two camera positions $\mathbf{t} = \mathbf{C} - \mathbf{C}^\prime$, which can be computed once per frame as it does not depend on the feature we are currently testing. Secondly we calculate the vector $\mathbf{p}$ as the projection in spherical coordinates of the image point $\mathbf{u}$, which is the position of the feature $\mathbf{P}$ as seen at the previous time step. In order to perform the calculations in the second time step, $\mathbf{p}$ has to be transformed according to the rotation between the two camera positions, which is $\mathbf{R} \mathbf{p}$. 

The epipolar plane can then be defined by the unit normal:
\begin{equation} \label{eqn::epiPlane}
\Pi: \mathbf{n}^\prime = \frac{\mathbf{p} \times \mathbf{e}^\prime} {|\mathbf{p}  \times \mathbf{e}^\prime|}
\end{equation}
(with $\times$ representing the standard vector/cross product) which also lies on the unit projection sphere, and is a pole of the great circle defined by the intersection of the epipolar plane with the unit projection sphere. $\mathbf{e}$ and $\mathbf{e}^\prime$ are the respective epipoles on the sphere, corresponding to unit vectors of $\mathbf{t}$ with opposite directions. We then calculate the vector $\mathbf{p}^\prime$, which is the projection of current position of the image feature point $\mathbf{u}^\prime$ as determined by the optical flow. 

If the tracked feature corresponds to a feature that is static in the world, or a feature that is in motion but on the epipolar plane, then $\mathbf{p}^\prime$ is co-planar with $\mathbf{e}^\prime$ and $\mathbf{p}$, i.e. lies on the epipolar plane $\Pi$. To check this restriction, the absolute value of the scalar product of $\mathbf{p}^\prime$ with $\mathbf{n}^\prime$ 
\begin{equation} \label{eqn::epiError}
\xi_e = | \mathbf{n}^\prime \cdot \mathbf{p}^\prime |
\end{equation}
(with $\cdot$ representing the standard scalar/inner product).
As we're dealing solely with unit vectors, the range of the $\xi_e$ will be in the range $[0,1]$, where $0$ will mean perfect co-planarity.

The fisheye epipolar constraint itself is not a perfect motion classifier, as it has a limitation that if the observed feature moves on (or near) the epipolar plane, it will be misclassified as a static feature.

\subsubsection{Note on the geodesic distance to epipolar plane}

It is perhaps a little bit more natural to think about the error on the surface of $\mathbf{S}^2$, as this is the projection surface, in place of the Euclidean error in $\mathbf{R}^3$. Therefore, here we quickly examine the error based on the squared geodesic distance in place of the squared Euclidean distance. Given that $\mathbf{S}^2$ is the unit sphere, the geodesic distance is the angle between the point $\mathbf{p}$ and the great circle defined by the intersection of the plane $\Pi$ with the unit sphere $\mathbf{S}^2$:
\begin{equation} \label{eqn:geodesic}
  \xi_{\mathbf{S}^2} = \left|\arcsin\left(\mathbf{n}^\prime\cdot\mathbf{p}^\prime\right)\right|
\end{equation}
where $\xi_{\mathbf{S}^2}$ is in this case the absolute angle between the plane $\Pi$ and $\mathbf{p}^\prime$. Noting that the $\xi_{\mathbf{S}^2} \approx \sin(\xi_{\mathbf{S}^2})$ for small values of $\xi_{\mathbf{S}^2}$, then we realise the same solution as in (\ref{eqn::epiError}).
\begin{align}
  \xi_{\mathbf{S}^2} \approx \sin(\xi_{\mathbf{S}^2})
  = \left|\mathbf{n}^\prime\cdot\mathbf{p}^\prime\right| = \xi_e
\end{align}
This small angle approximation is valid, as even if the point $\mathbf{P}$ is dynamic and moves away from the epipolar plane $\Pi$, it will typically not be by a large distance. Even in the absence of the small angle assumption, $\xi_e$ is a valid measure of error, as both $\xi_e = \sin(\xi_{\mathbf{S}^2})$, and if $\theta_1 > \theta_2$ are any arbitrary values in the range $\left[0, \frac{\pi}{2}\right]$, then $\sin(\theta_1) > \sin(\theta_2)$. That is, as $\xi_{\mathbf{S}^2}$ increases monotonically with distance of $\mathbf{P}$ from the plane, so does $\xi_e$.

\subsubsection{Note on odometry scale} The epipolar constraint (both the pinhole and the planar) are independent of absolute odometry scale. In the former, $\textbf{E}$ describes the translation and rotation between the camera pair, with $\mathbf{t}$ typically known only up to scale, but the constraint still applies. In the planar case, it can be seen from (\ref{eqn::epiPlane}) and (\ref{eqn::epiError}) that the epipolar error is based only on the epipole $\mathbf{e}^\prime$, which is a unit vector form of $\mathbf{t}$, i.e. the scale of $\mathbf{t}$ is unimportant here. This is geometrically intuitive as well. Observing Figure \ref{fig:PlanarEpipolar}, it can be seen the distance from the point $\mathbf{p}^\prime$ to the plane $\Pi$ is independent of the overall scale of the system. This is useful, as it can be employed directly in systems in which the odometry is known only up to scale, such as visual odometry and visual SLAM systems.

\subsection{Fisheye Positive Depth Constraint}
The {\em positive depth} or {\em cheirality constraint} requires all imaged points to lie ``in front'' of the camera, and solves a class of feature motion that is not solved by the fisheye epipolar constraint. However, for a fisheye image, the term ``in front'' is not well defined, as the rays can in fact point behind (negative $Z$ in camera coordinates) the cameras. To be more definite, if we interpret the rays as lines, their parametric equations are 
\begin{align}
\mathcal{L}: \mathbf{x} & = \mathbf{C} + t \mathbf{p} \nonumber \\
\mathcal{L}^\prime: \mathbf{x}^\prime & = \mathbf{C}^\prime + t^\prime \mathbf{p}^\prime
\label{eqn:midpoint}
\end{align}
If the point $\mathbf{P}$ is unmoving, then their convergence point (the point at which each line passes closest to one another) occurs when both $t$ and $t^\prime$ are positive, thus describing a positive depth along the lines towards the point of convergence. finding the closest point on $\mathcal{L}$ and $\mathcal{L}^\prime$ is the basis of the {\em midpoint method} of reconstruction. While the midpoint method is provably non-optimal method of reconstruction (particularly in projective reconstruction) \cite{hartley1997}, it is still often used due to its simplicity and low computational cost. Its implementation will typically be done by constructing the line equations (\ref{eqn:midpoint}) and finding the midpoint of the line between the two closest points on $\mathcal{L}$ and $\mathcal{L}^\prime$. In such a case, the positive depth constraint can be implemented simply by analysing the signs of $t$ and $t^\prime$ at the closest passing of $\mathcal{L}$ and $\mathcal{L}^\prime$, and if one or both are negative, then this indicates the likelihood of $\mathbf{P}$ being under motion.

Another interpretation of this is that, if we consider the points $\mathbf{p}$ and $\mathbf{p}^\prime$ on the unit sphere $\mathbf{S}^2$, then the directional arc on the sphere formed by $\mathbf{p}$ and $\mathbf{p}^\prime$ must point towards $\mathbf{e}^\prime$. That is, if the following is true
\begin{align}
\mathbf{n}^\prime \cdot \mathbf{p}^\prime & = 0 \label{eqn:PosDepthSphere1} \\
\mathbf{p}^\prime \cdot \mathbf{e}^\prime & < \mathbf{p} \cdot \mathbf{e}^\prime \label{eqn:PosDepthSphere2}
\end{align}
then the point can be considered moving. (\ref{eqn:PosDepthSphere1}) is the restriction that if $\mathbf{P}$ is static, then $\mathbf{p}$, $\mathbf{p}^\prime$ and $\mathbf{e}^\prime$ must be on the same great circle of $\mathbf{S}^2$, and is basically the epipolar constraint from (\ref{eqn::epiError}). (\ref{eqn:PosDepthSphere2}) is the restriction that $\mathbf{p}^\prime$ must be closer to $\mathbf{e}^\prime$ than $\mathbf{p}$ on the same great circle if point $\mathbf{P}$ is static. However, this formulation suffers from a subtle restriction: it only applies in the case that the $\mathbf{P}$ is moving but $\mathbf{p}^\prime$ remains on the epipolar plane (i.e. the fisheye epipolar check fails to identify the point). We prefer an approach that uses the same principles, but can be applied to all points.

\begin{figure}[tb]
  \centering
  \includegraphics[trim={6cm 3cm 5cm 7.5cm},clip,width=\columnwidth]{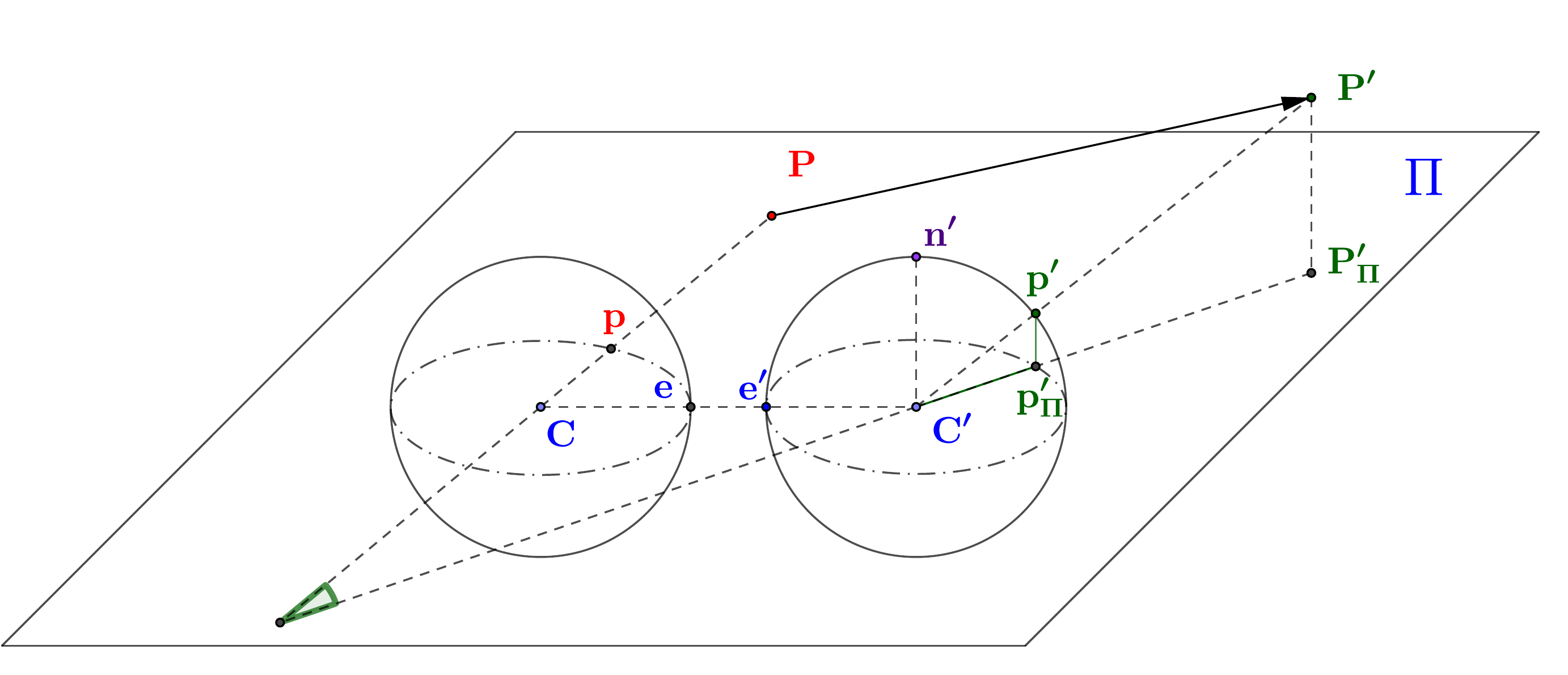}
  \vspace{-0.6cm}
  \caption{The positive depth constraint. The camera moved from $\mathbf{C}$ to $\mathbf{C}^\prime$ and the observed feature point moved from $\mathbf{P}$ to $\mathbf{P}^\prime$. The epipolar plane $\Pi$ is defined by the three points $\mathbf{C}$, $\mathbf{C}^\prime$ and $\mathbf{P}$. The vectors $\mathbf{p}$ and $\mathbf{p}^\prime$ are the projections of $\mathbf{P}$ and $\mathbf{P}^\prime$ on the unit spheres at the respective time steps. The rays pointing to $\mathbf{P}$ and $\mathbf{{P}^\prime_{\Pi}}$, which is the projection of $\mathbf{P}^\prime$ on the epipolar plane, converge behind the camera position, which means that the positive depth constraint is not satisfied. }
  \label{fig:posDepth}
  \vspace{-0.4cm}
\end{figure}

With reference to Figure \ref{fig:posDepth}, if the rays from $\mathbf{P}$ and $\mathbf{P}^\prime$, or their corresponding unit sphere ray $\mathbf{p}$ and $\mathbf{p}^\prime$, do not converge ``in front'' of the camera, then the point can be considered moving. This can be checked by first considering the unit vector of the projection of $\mathbf{p}^\prime$ on the epipolar plane, given by 
\begin{equation}
\mathbf{p}^\prime_{\Pi} = \mathbf{p}^\prime - (\mathbf{p}^\prime \cdot \mathbf{n}^\prime)\mathbf{n}^\prime
\label{eqn:posDepthProjEpi}
\end{equation}
as is shown in Figure \ref{fig:PlanarEpipolar}. Note that $\mathbf{p}^\prime$ can be on the epipolar plane $\Pi$, in which case $\mathbf{p}^\prime_{\Pi} = \mathbf{p}^\prime$. Utilising the vector product
\begin{equation}
\mathbf{p_n} = \mathbf{p}^\prime_{\Pi} \times \mathbf{p}
\label{eqn:posDepthCross}
\end{equation}
returns a vector that is orthogonal to the epipolar plane, but may be in the same direction as the previously defined epipolar plane normal $\mathbf{n}^\prime$, may be in the opposite direction (relative to the epipolar plane), or may be the zero vector. The directionality can be checked using the scalar product. That is, if:

\begin{itemize}
    \item {$\mathbf{n}^\prime \cdot \mathbf{p_n} < 0$}: the vectors $\mathbf{n}^\prime$ and $\mathbf{p_n}$ lie in the same direction, and $\mathbf{p}^\prime$ and $\mathbf{p}$ converge in front of the camera
    \item {$\mathbf{n}^\prime \cdot \mathbf{p_n} > 0$}: the vectors $\mathbf{n}^\prime$ and $\mathbf{p_n}$ lie in opposite directions, and $\mathbf{p}^\prime$ and $\mathbf{p}$ converge behind the camera
    \item {$\mathbf{n}^\prime \cdot \mathbf{p_n} = 0$}: $\mathbf{p_n}$ is the zero vector, and $\mathbf{p}^\prime$ and $\mathbf{p}$ do not converge (they are parallel)
\end{itemize}
Therefore, we can define the positive depth constraint as
\begin{equation}
\xi_d =\left\{
                \begin{array}{ll}
                  |\mathbf{p_n}|, & \mathbf{n}^\prime \cdot \mathbf{p_n} > \ 0 \\
                  0, & \text{otherwise}
                \end{array}
              \right.
\label{eq:posDepth}
\end{equation}
$\xi_d$ is the the sine of the angle between $\mathbf{p}^\prime_{\Pi}$ and $\mathbf{p}$, and is in the range $[0, 1]$ since all vectors are unit vectors. 

Like the epipolar constraint, the fisheye positive depth constraint does not require the relative positions of the cameras $\mathbf{C}$ and $\mathbf{C}^\prime$ (i.e. odometry) with scale. This can be seen by the fact that the equations that define the positive depth constraint ((\ref{eqn:posDepthProjEpi}) and (\ref{eqn:posDepthCross})) consists solely of operations on unit vectors. It can also be understood geometrically -- the absolute distance between the two camera positions $\mathbf{C}$ and $\mathbf{C}^\prime$ does not affect the angle at the convergence point. 

The positive depth constraint can detect when a feature's motion projected to the epipolar plane is greater than the movement of the camera itself. Roughly speaking, in the vehicle context, this will detect when the other obstacle is moving faster than the host vehicle, in the same direction as the host vehicle (for example, overtaking vehicles, which fail the fisheye epipolar check).

\subsection{Fisheye Positive Height Constraint}

The {\em fisheye positive height constraint} is the first of the constraints 
that impose a specific structure on the scene - that a road plane exists, and that its location relative to the camera is approximately known. In such a case, it is reasonable to assume that
if feature vectors converge below the road plane, we can consider them to be moving in the scene. Inaccuracies in this assumption are handled through the use of a threshold. We assume we know the height of the camera from the road plane, $\eta_{C}$, and the rotation of the camera with respect to the road plane, $\mathbf{R}_{C}$.
The calibration of cameras relative to the road is well known \cite{schmidt2018}
, and it is a fair assumption that such a calibration exists in an automotive system.

The positive height constraint applies only if the observed point in the image at both the previous ($\mathbf{u}$) and current positions ($\mathbf{u}^\prime$) are below the horizon line corresponding to the road plane. The equivalent statement for the spherical coordinates is that the vectors $\mathbf{p}$ and $\mathbf{p}^\prime$ must be below the plane through the camera centre $\mathbf{C}^\prime$ parallel to the road plane. The vector defining the horizon plane in camera coordinates $\mathbf{h}$ is therefore the vector perpendicular to the ground plane in world coordinates, pointing downwards, multiplied by the rotation matrix $\mathbf{R}_{C}$.
\begin{equation} \label{eq:horizon}
\mathbf{h} = \mathbf{R}_{C} [0, 0, -1]^\top
\end{equation}
Since $\mathbf{h}$ points downwards, the conditions to be met are $\mathbf{p}^\prime \cdot \mathbf{h} > 0$ and $\mathbf{p} \cdot \mathbf{h} > 0$. 

\begin{figure}[tb]
  \centering
  \includegraphics[trim={6cm 1.8cm 4.5cm 8cm},clip,width=0.8\columnwidth]{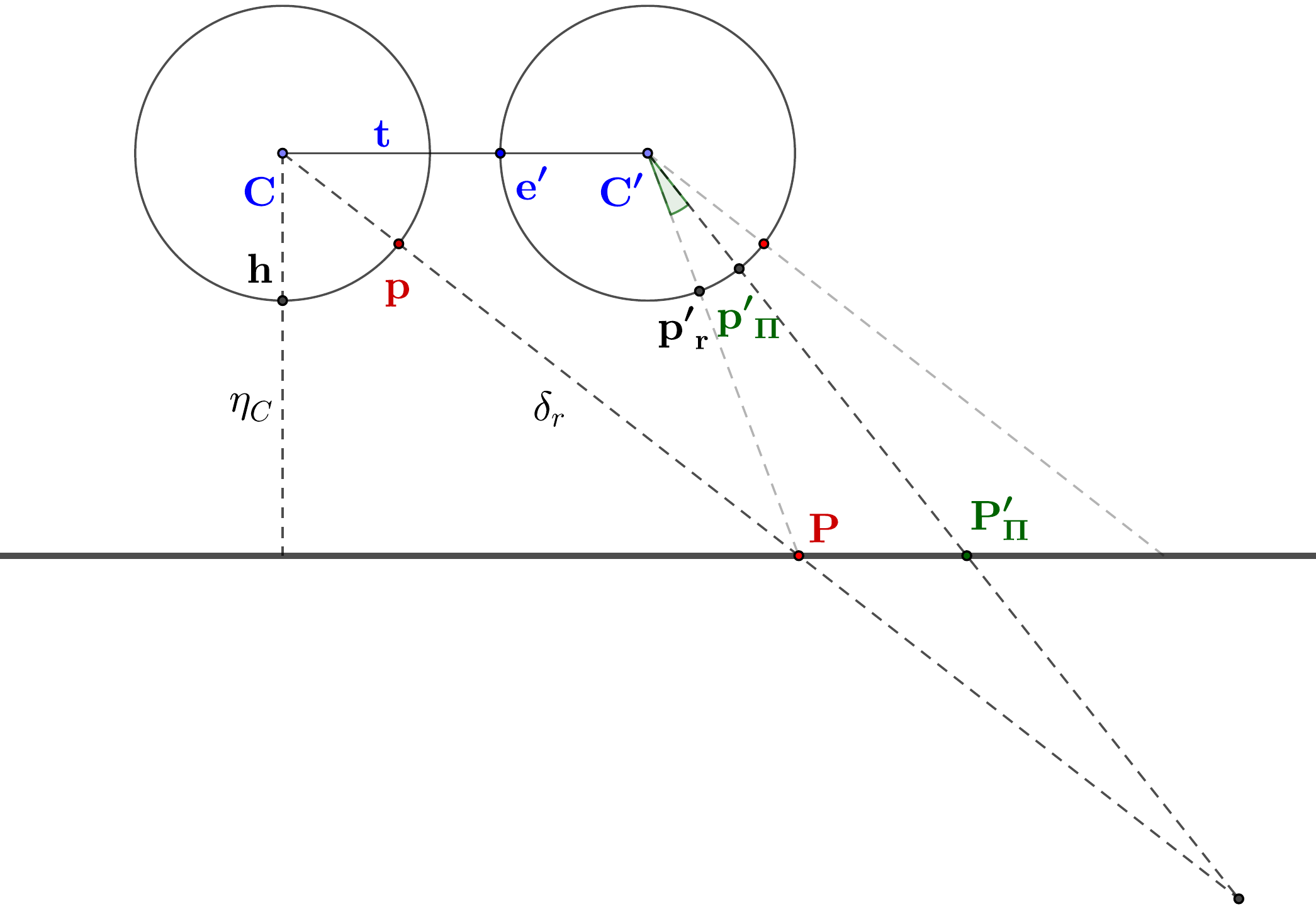}
  \vspace{-0.3cm}
  \caption{The positive height constraint, viewed as a projection onto the epipolar plane. The page can be considered the epipolar plane through the camera positions $\mathbf{C}$, $\mathbf{C}^\prime$ and the feature point $\mathbf{P}$, referred to as $\mathbf{\Pi}$. The circles are the intersections of the unit spheres with the epipolar plane. $\mathbf{P}^\prime_{\Pi}$ and $\mathbf{p}^\prime_{\Pi}$ are the projection of $\mathbf{P}^\prime$ and $\mathbf{p}^\prime$ respectively. The rays pointing to $\mathbf{P}$ and $\mathbf{{P}^\prime_{\Pi}}$ converge below the ground plane (bold black line), which means that the positive height constraint is not satisfied.}
  \label{fig:posHeight}
  \vspace{-0.4cm}
\end{figure}

With reference to Figure \ref{fig:posHeight}, the vector $\mathbf{p}^\prime_{r}$ is the point on the unit sphere that corresponds to the intersection of the previous point vector $\mathbf{p}$ with the road plane, represented by $\mathbf{P}$ on the road plane, in the current spherical coordinate system.
\begin{equation}
\mathbf{p}^\prime_{r} =  (\delta_{r} \cdot \mathbf{p}) + \mathbf{t}
\end{equation}
The distance $\delta_{r}$ can be calculated defining a triangle with sides $\eta_{C} \cdot \mathbf{h}$  (vertical from the camera to the road), the direction of $\mathbf{p}$ and the road plane. As the cosine of the angle between $\mathbf{p}$ and $\mathbf{h}$ is $\mathbf{p} \cdot \mathbf{h}$ :
\begin{equation}
\delta_{r} = \frac{\eta_{C}}{\mathbf{p} \cdot \mathbf{h}}
\end{equation}
The rays through $\mathbf{P}$ and $\mathbf{P}^\prime_{\Pi}$ are below the horizon and cross below the road plane if $\mathbf{p}^\prime_{\Pi}$ is between $\mathbf{p}$ and $\mathbf{p}^\prime_{r}$. The two conditions are respectively met if $\mathbf{n}^\prime \cdot \mathbf{p_n} < 0$ and $\mathbf{n}^\prime \cdot  \mathbf{p} > 0$.

If these two conditions are met, the positive height deviation is the length of the vector $\mathbf{v}$, where
\begin{equation}
\mathbf{v} = \mathbf{p}^\prime_{\Pi} \times \mathbf{p}^\prime_{r}
\end{equation}
The positive height constraint is therefore
\begin{equation}
\xi_h =\left\{
            \begin{array}{ll}
              |\mathbf{v}| - \lambda_h, & \mathbf{p}^\prime \cdot \mathbf{h} > 0 \text{ and } \mathbf{p} \cdot \mathbf{h} > 0 \text{ and } \\ & \mathbf{n}^\prime \cdot \mathbf{p_n} < 0 \text{ and } \mathbf{n}^\prime \cdot  \mathbf{p} > 0 \\
              0, & \text{otherwise}
            \end{array} 
          \right.
\end{equation}
where the threshold $\lambda_h$ was set to 0.001, which was found as a good value to suppress bad detections due to noise after an empirical analysis of the scenes presented in the Section \ref{sec:results}.

The positive height constraint requires knowledge of odometry with scale resolution, as this is require to determine if the triangulation is below the road surface.

\subsection{Anti-parallel Constraint}

A category of moving objects that is going to be missed from the previous classification is the one of objects whose motion mirrors the ego-vehicle, which we are going to refer to as {\em anti-parallel}. 
For example, this poses a problem in the detection of approaching vehicles in the opposite lane, since this is a common situation in road scenarios.

\begin{figure}[tb]
  \vspace{-0.4cm}
  \centering
  \includegraphics[trim={12cm 6.5cm 9cm 6cm}, clip,width=0.8\columnwidth]{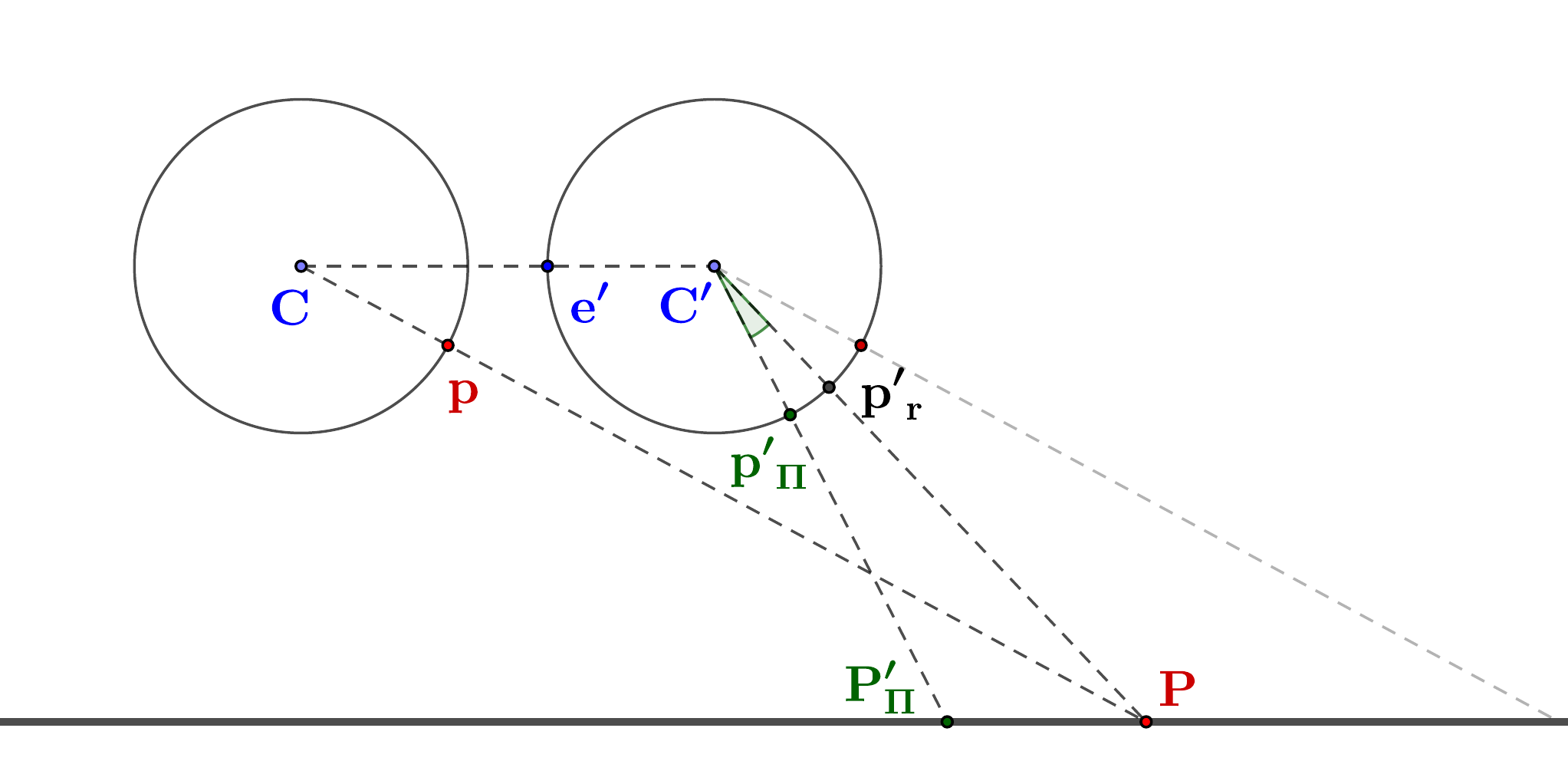}
  \vspace{-0.3cm}
  \caption{The anti-parallel constraint calculation, viewed as a projection onto the epipolar plane. The page can be considered the epipolar plane through the camera positions $\mathbf{C}$, $\mathbf{C}^\prime$ and the feature point $\mathbf{P}$, referred to as $\mathbf{\Pi}$. The circles are the intersections of the unit spheres with the epipolar plane. $\mathbf{P}^\prime_{\Pi}$ and $\mathbf{p}^\prime_{\Pi}$ are the projection of $\mathbf{P}^\prime$ and $\mathbf{p}^\prime$ respectively. The rays pointing to $\mathbf{P}$ and $\mathbf{{P}^\prime_{\Pi}}$ converge above the ground plane (bold black line). If the angle between $\mathbf{p}^\prime_{r}$ and $\mathbf{p}^\prime_{\Pi}$ is greater than the defined threshold, then we consider the anti-parallel constraint to be not satisfied.}
  \label{fig:antiparallel}
  \vspace{-0.5cm}
\end{figure}

Referring to the Figure \ref{fig:antiparallel}, in this case we reason in the opposite way to the positive height constraint. If the vector $\mathbf{p}^\prime_{\Pi}$ is below the horizon and behind $\mathbf{p}^\prime_{r}$ (given $\mathbf{n}^\prime \cdot \mathbf{p_n} < 0$ and $\mathbf{n}^\prime \cdot \mathbf{p} < 0$), then the point triangulates above the road plane, as shown in Figure \ref{fig:antiparallel} and could correspond either to a static object or to an approaching object. To differentiate between the two cases, we introduce a threshold value $\lambda_p$. If the angle between $\mathbf{p}^\prime_{r}$ and $\mathbf{p}^\prime_{\Pi}$ is greater than the threshold, then the difference 
\begin{equation}
\xi_h =\left\{
            \begin{array}{ll}
              |\mathbf{v}| - \lambda_p, &  |\mathbf{v}| > \lambda_p \\
              0, & \text{otherwise}
            \end{array} 
          \right.
\end{equation}
is defined as the anti-parallel constraint. The value of the threshold $\lambda_p$ can be set as a constant. In our case, a value of 0.001 was appropriate. 

The anti-parallel constraint requires knowledge of odometry with scale resolution, we essentially need to predict the length of flow on the image sphere for a point on the road surface, which requires knowledge of the absolute translation of the camera.

\subsection{Three-view Constraint}

\begin{figure}[tbh]
  \vspace{0cm}
  \centering
  \includegraphics[trim={6cm 1cm 4cm 1cm}, clip,width=0.8\columnwidth]{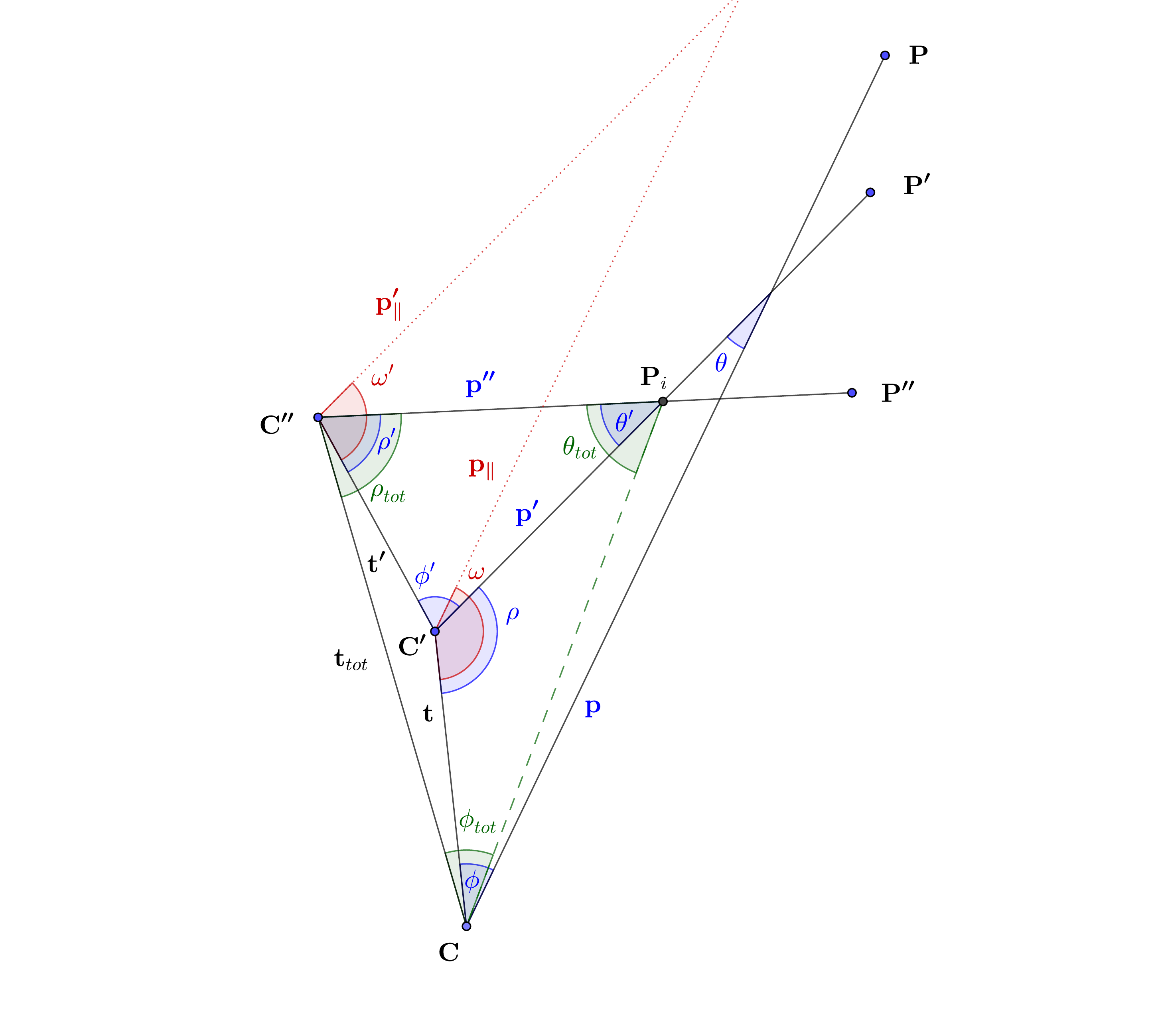}
  \vspace{-0.6cm}
  \caption{The three-view constraint (the page can be considered the epipolar plane). The continuous lines connect the camera positions to the point position observed at the three time steps (unit vectors $\mathbf{p}$, $\mathbf{p}^\prime$, $\mathbf{p}^\prime{}^\prime$), the dashed lines are the direction in the current view of the point as seen in the previous view (unit vectors $\mathbf{p_{\parallel}}$, $\mathbf{p}^\prime_{\parallel}$). All the vectors are the projection on the epipolar plane calculated on the first two frames. The subscript $\Pi$ and the unit spheres were removed to ease the readability of the image.}
  \label{fig:three_view}
  \vspace{-0.5cm}
\end{figure}

The use of {\em three view constraint} \cite{hartley1997_tri} 
for motion segmentation is reasonably well known \cite{torr1997, klappstein2009}. The processing of three views adds complexity in that it requires a triple of points to be matched between frames, i.e. $\mathbf{u} \leftrightarrow \mathbf{u}^\prime \leftrightarrow \mathbf{u}^\prime{}^\prime$, which is not always available. Typically, dense optical flow cannot retrack points, so cannot provide correspondences over three frames. Sparse or feature based approaches can be configured to provide such correspondences. The previously cited work uses the trifocal tensor for the calculation of the trifocal error over three images. Since here we are using points on the unit sphere $\mathbf{p} \leftrightarrow \mathbf{p}^\prime \leftrightarrow \mathbf{p}^\prime{}^\prime$, the approach has to be modified. Figure \ref{fig:three_view} shows a representation of the triangulation of a generic moving point over three frames. 

We build two triangles with the translation vectors ($\mathbf{t}$ and $\mathbf{t}^\prime$), for which knowledge of the absolute scale is required, and the intersections of the connecting lines. A third triangle can be built with base the sum of the two translation vectors ($\mathbf{t}_{tot}$). If the point is static, the sum of the angles opposite to the translation vectors in the two partial triangles ($\theta$ and $\theta^\prime$) will equal the angle of the total triangle ($\theta_{tot}$). If the angles have different values, the sine of the difference between the sum $\theta$ + $\theta^\prime$ and $\theta_{tot}$ is defined as the three view constraint.

In order to calculate $\theta$, we calculate the values of $\omega = \angle(\mathbf{p}_{\parallel}, \mathbf{t})$ and $\rho = \angle(\mathbf{p}^\prime, \mathbf{t})$. Since $\mathbf{p}_{\parallel}$ and $\mathbf{p}$ are parallel lines, $\theta = \omega - \rho$. The same applies to the triangle based on $\mathbf{t}^\prime$. In order to find $\theta_{tot}$, we use the fact that the triangles built on $\mathbf{t}_{tot}$ and $\mathbf{t}^\prime_{tot}$ share one side, $\mathbf{C}^\prime{}^\prime \mathbf{P}_i$, where $\mathbf{P}_i$ is the intersection between the directions of the vectors $\mathbf{p}^\prime$ and $\mathbf{p}^\prime{}^\prime$. The common side $\mathbf{C}^\prime{}^\prime \mathbf{P}_i$ can be written as
\begin{equation}
\mathbf{C}^\prime{}^\prime \mathbf{P}_i = \mathbf{t}^\prime \cdot \frac{\sin(\phi^\prime)}{\sin(\theta^\prime)}
\end{equation}
Using the law of cosines, the third side of the $\mathbf{t}_{tot}$ triangle, $\mathbf{C} \mathbf{P}_i$, can be written as
\begin{equation}
\mathbf{C} \mathbf{P}_i = \sqrt{\mathbf{t}_{tot}^2 + \mathbf{C}^\prime{}^\prime \mathbf{P}_i^2 - 2 \cdot \mathbf{t}_{tot} \cdot \mathbf{C}^\prime{}^\prime \mathbf{P}_i \cdot \cos(\rho_{tot})}
\end{equation}
Finally the angle $\theta_{tot}$ can be find again by the law of sines as 
\begin{equation}
\theta_{tot} = \sin^{-1} \left( \sin(\rho_{tot}) \cdot \frac{\mathbf{t}_{tot}}{\mathbf{C} \mathbf{P}_i} \right)
\end{equation}
The final deviation from the three view constraint is calculated as the sine of the absolute difference in the angles at the intersection point $\mathbf{P}_i$, that is 
\begin{equation}
\xi_{3v} = \sin(|(\theta + \theta^\prime) - \theta_{tot}|)
\end{equation}

\subsection{Static Camera Degenerate Case}

For completeness, we include here also the degenerate case where the camera itself is still. The four other constraints don't work in this case, as $\mathbf{C}$ and $\mathbf{C}^\prime$ coincide and $\mathbf{e}^\prime$ cannot be defined. In this case, we define the geometric constraint deviation as:
\begin{equation}
\label{eq:degen}
\xi = |\mathbf{p}^\prime \times \mathbf{p}| 
\end{equation}
which is simply a measure of optical flow once projected to the unit sphere.

There is the possibility that if the camera is starting to move from a static position, the position is not updated accordingly until the movement is sufficiently big, e.g. if the odometry is calculated through mechanical means such as the ticks of the wheels. If the position has not been updated, the calculation of $\xi$ will be performed under the static degenerate case even if, in fact, the camera has started moving. This motion will usually produce flow vectors on the road plane, which can potentially be detected as valid moving objects. In the absence of tracking over more than two frames, it is not possible to know if the camera position will be updated in the next frames. 

To avoid these systematic false positives, a criterion can be introduced for the feature vectors below the horizon. Similarly to the positive height constraint, both $\mathbf{p}$ and $\mathbf{p}^\prime$ can be projected on the road plane by multiplying them for the distances $\delta_r$ and $\delta_r^\prime$ calculated as 

\begin{equation}
    \label{eq:degen2}
    \delta_r^\prime = \frac{\eta_C}{\mathbf{p}^\prime \cdot \mathbf{h}}
\end{equation}

The displacement on the road plane $\mathbf{d}$ is calculated as the difference $\mathbf{d} = \mathbf{P}^\prime - \mathbf{P}$. 
If the magnitude of $\mathbf{d}$ is smaller than a threshold $\lambda_s$, $\xi$ is set to 0

\begin{equation}
\xi =\left\{
                \begin{array}{ll}
                  0, &  |\mathbf{d}| < \lambda_s    \\
                  |\mathbf{p}^\prime \times \mathbf{p}|, &  \text{otherwise} \\
                \end{array} 
              \right.
\end{equation}
Since for the camera to be still considered as static the actual motion between frames is going to be small (i.e. in the centimeter range), setting the threshold in this range assures that real moving objects will not be filtered out, as the projection of real moving points on the road are likely to generate a much bigger displacement.

\subsection{Motion Likelihood Calculation}
After the individual deviation components are calculated for a point in the image for each constraint ($\xi_e$, $\xi_d$, $\xi_h$, $\xi_p$ and possibly $\xi_{3v}$), they are combined into a metric that quantifies the likelihood that the point is moving rather than static, i.e. motion likelihood. The final motion likelihood is calculated as the weighted mean of the four individual deviations (ignoring the simple degenerate case)
\begin{equation}
    \xi = \frac{\sum_{i}^{} \mu_i \xi_i}{\sum_{i}^{} \mu_i}
\end{equation}
where $i \in \{e, d, h, p, 3v\}$ and ($\mu_e$, $\mu_d$, $\mu_h$, $\mu_p$, $\mu_{3v}$) are the weights assigned to the constraint deviation components. In the case that the three-view constraint is not possible (due to retracked optical flow inavailability, for example), $\mu_{3v} = 0$. In our case, the values of the weights were empirically set to (1.0, 1.0, 0.2, 0.2, 0) in order to assign more importance to the epipolar and positive depth constraints, which are always true, as opposed to the positive height and anti-parallel ones that require stronger assumptions on the scene. 

\begin{figure}[hbt]
  \vspace{0cm}
  \centering
  \includegraphics[trim={5cm 3cm 6cm 0cm}, clip,width=0.7\columnwidth]{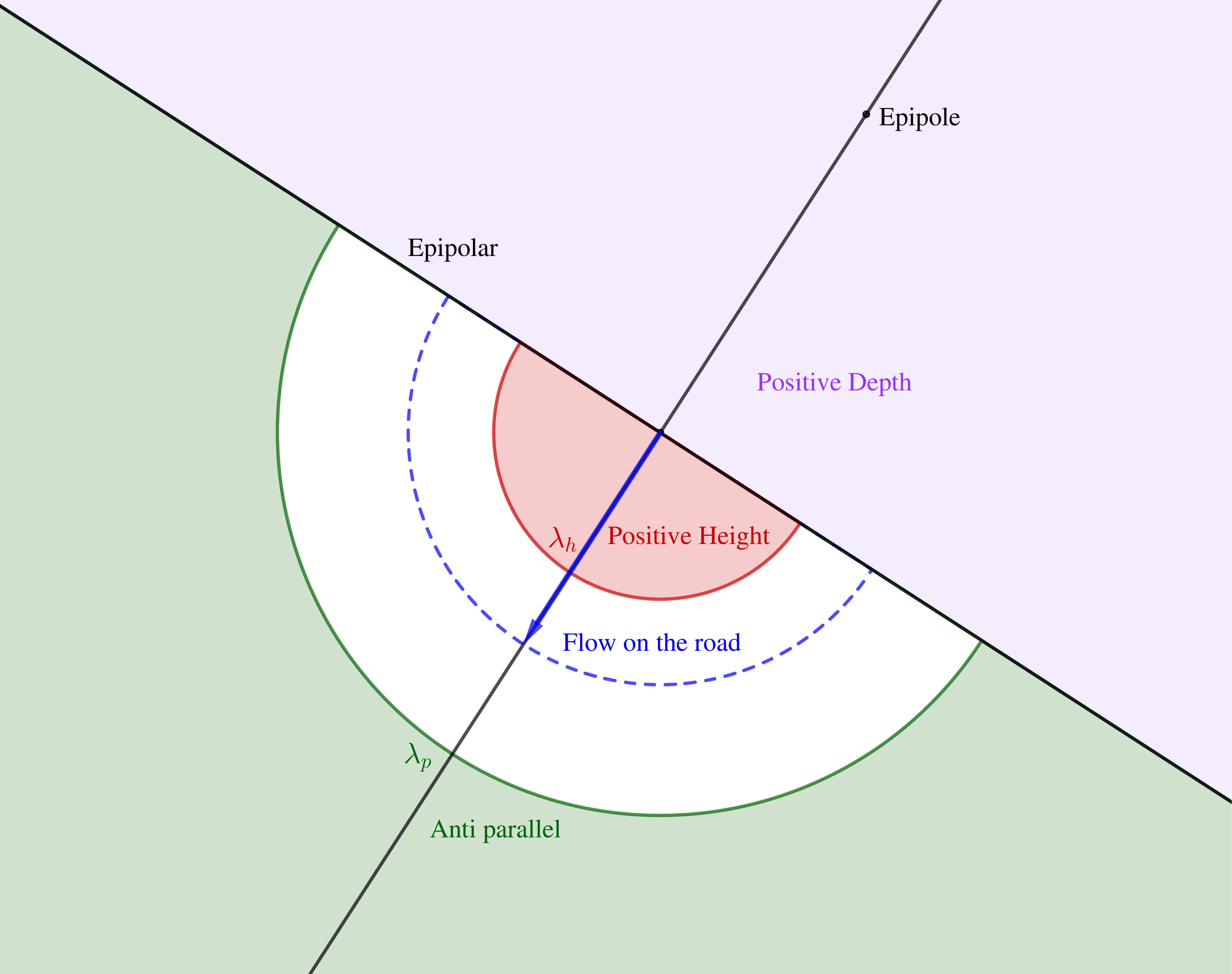}
  \caption{This diagram represents the areas of the image plane where the constraints apply, given the epipole and the optical flow vector expected on the road plane (blue vector). The flow on the road is pointing in the direction opposite to the epipole. The epipolar constraint is measured as the component perpendicular to the direction going through the epipole. The positive depth constraint applies to the area opposite the general direction of the flow in the image. Positive height and anti parallel constraints apply on the same direction of the flow on the road in the areas determined by the thresholds used. }
  \vspace{-0.5cm}
  \label{fig:thresholds}
\end{figure}

Figure \ref{fig:thresholds} shows a graphical representation of the areas of the image where the individual constraints apply, given the position of the epipole and of the flow on the road plane. For each feature (or pixel in the case of dense optical flow), $\xi$ is thresholded to provide the motion segmentation output (e.g. which thresholds row VI of Figure \ref{fig6} into row VII). The exact value for the threshold guides the overall sensitivity of the motion segmentation.

\section{Results}

\label{sec:results}

In this section we present results for different categories of moving objects. The dense optical flow was calculated using the Farneb{\"a}ck algorithm \cite{farneback2003} on the frames of size $640 \times 480$ pixels, and averaged into $5 \times 5$ cells. 

\subsection{Subjective Analysis}
\begin{figure*}[t]
  \centering
  \includegraphics[trim={3cm 17.5cm 1.5cm 0.5cm}, clip,width=\textwidth]{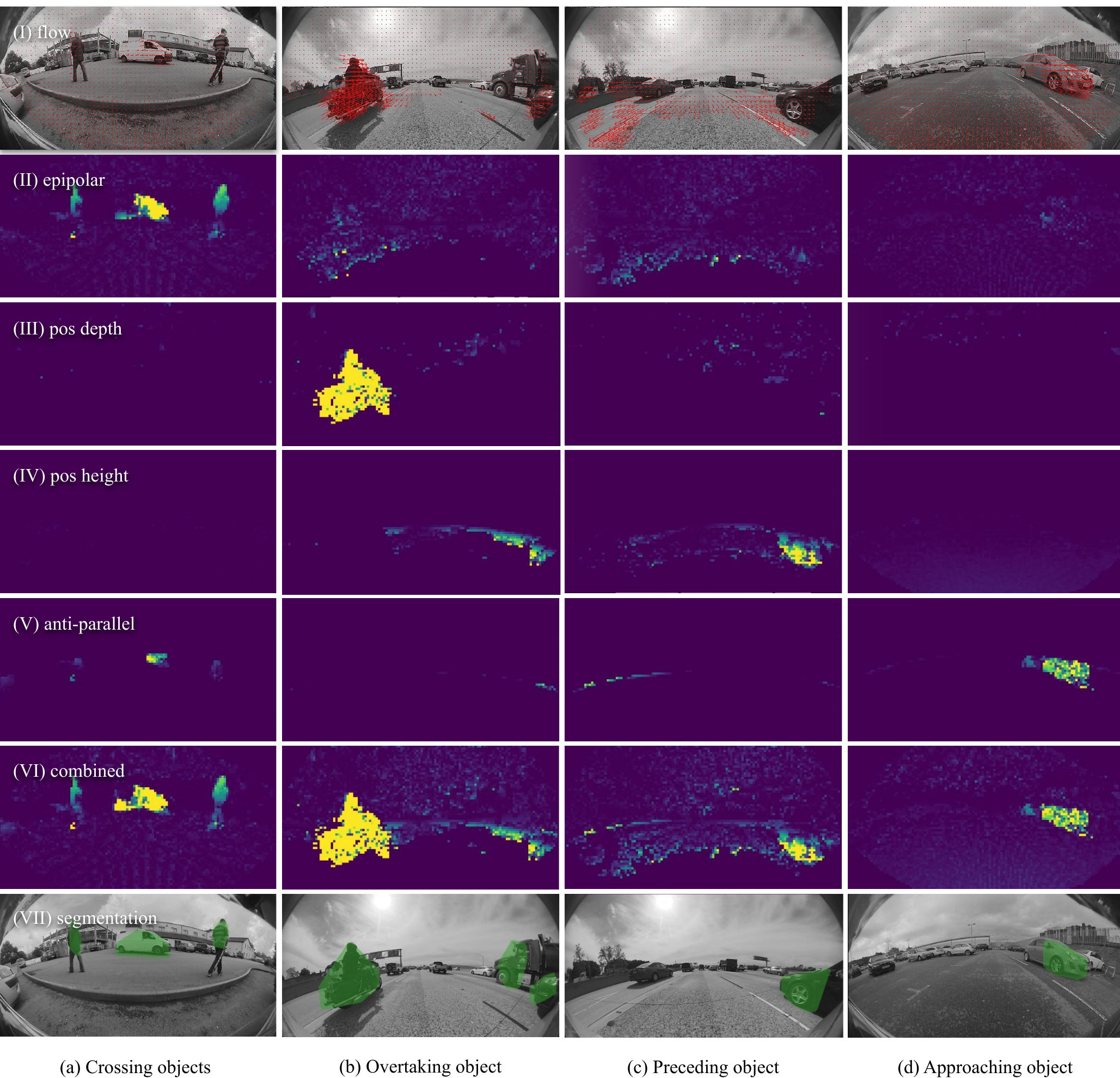}
  \vspace{-0.7cm}
  \caption{Frames with representative results for each type of detectable moving object. Row (I) shows the flow vectors. Rows (II)-(V) show the deviation measured by the individual constraints. Row (VI) shows the final motion likelihood resulting from the average of the four constraint components. Theoretically, the range of motion likelihood is $[0, 1]$, but in practice they rarely exceed 0.02. The colour map is therefore saturated at 0.02. The last row (VII) shows the final motion segmentation obtained by thresholding $\xi$ with a value $6 \times 10^{-4}$. }
  \vspace{-0.35cm}
  \label{fig6}
\end{figure*}

\begin{figure}[!tbp]
  \centering
  \includegraphics[trim={0.7cm 10.5cm 0.7cm 0.5cm}, clip,width=\columnwidth]{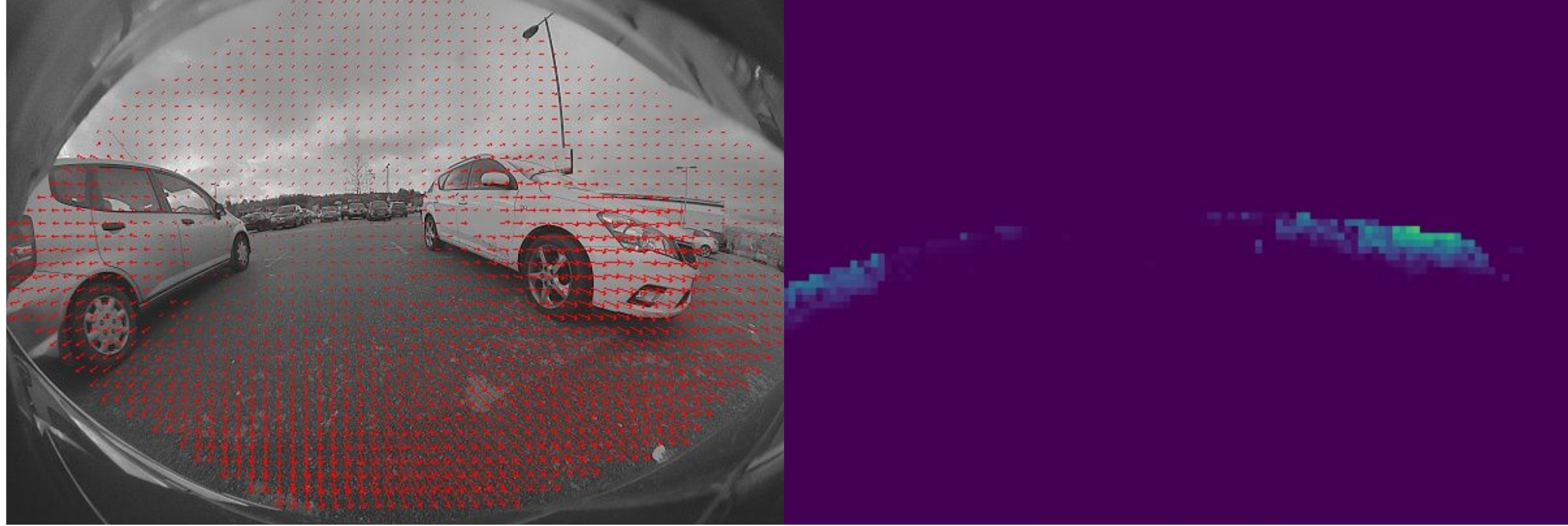}
  \vspace{-0.7cm}
  \caption{False positive detections caused by the anti-parallel constraint. Both cars are static, but a response from the anti-parallel constraint can be seen.}
  \label{fig7}
  \vspace{+0.5cm}
\end{figure}

A visualization of the results is included in Figure \ref{fig6}. 
The figure shows the results for the types of motion described in Figure \ref{fig::little}. It can be seen that each of the geometric motion segmentation approaches solves the detection problem for given types of motion, and in combination, all types of object motion are detected. As observed in Figure \ref{fig6}(c-d), the positive height and anti-parallel constraints are only valid below the horizon in the image. A drawback of the anti-parallel constraint is that its criteria are satisfied also by static objects high on the ground and/or close to the camera (Figure \ref{fig7}), causing false positives on close static objects. This is the source of most of the false positives in Table \ref{table_2}. 
The work presented in \cite{mariotti2019} contains a deeper analysis of this figure and of the visual performance of the constraints.

\begin{figure}[!tbp]
  \centering
  \includegraphics[trim={1cm 8.5cm 1.7cm 1cm}, clip,width=\columnwidth]{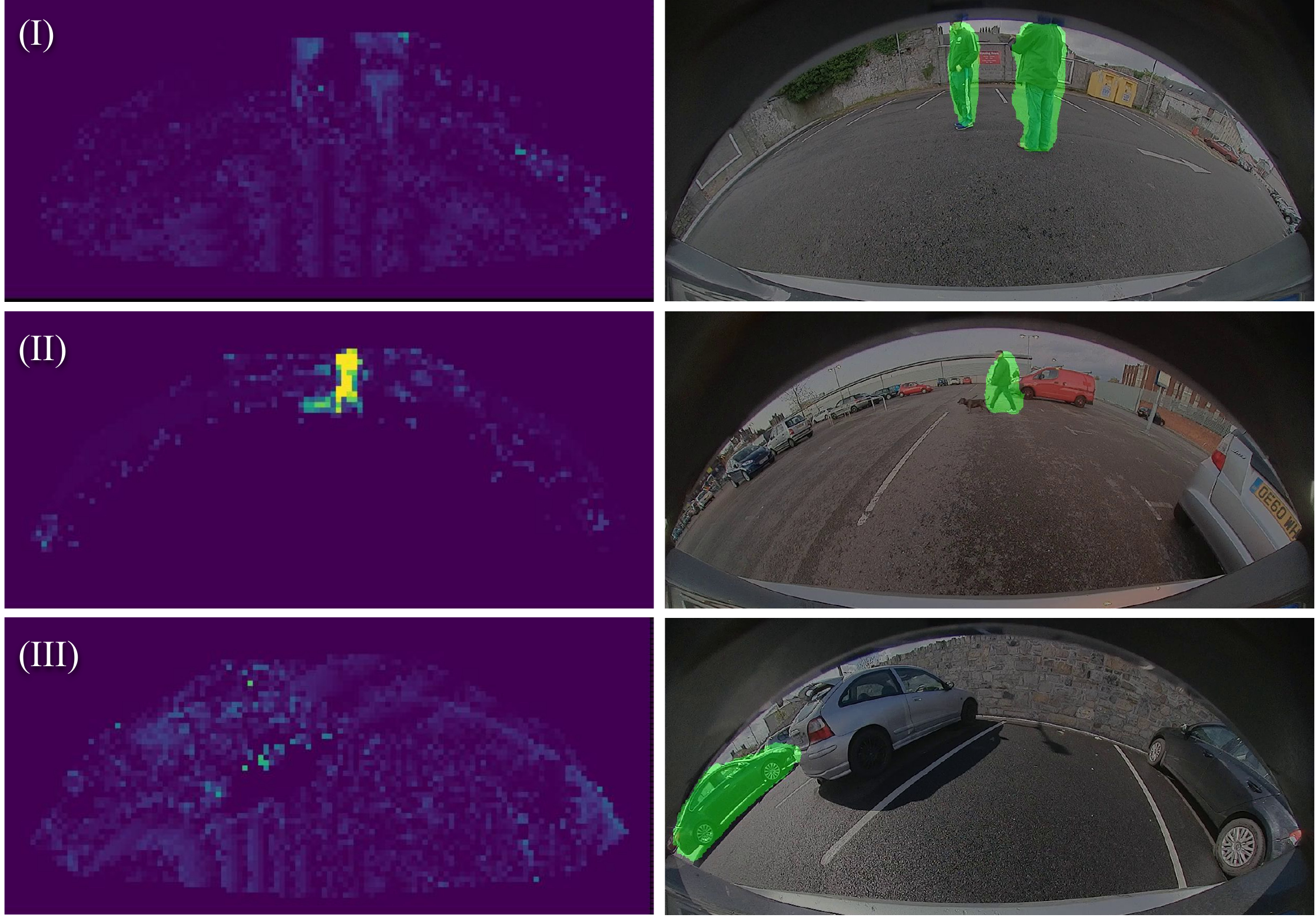}
  \caption{Qualitative comparison between our method (left) and FisheyeMODNet \cite{yahiaoui2019} (right). Row I: Two static pedestrians, falsely detected by FisheyeMODNet as moving. Row II: Correctly detected moving pedestrian by both methods. Row III: Moving vehicle missed by our method.}
  \label{fig8}
  \vspace{-0.5cm}
\end{figure}

Recent work (FisheyeMODNet \cite{yahiaoui2019}) performed training on a fisheye data set, which showed improvement of the performance on fisheye images over a standard camera trained network 
\cite{siam2018, wang2018}. In addition to what presented in \cite{mariotti2019}, in Figure \ref{fig8} we provide a qualitative comparison of how some specific objects are detected by our method and by FisheyeMODNet. In row (I) two static pedestrians are correctly not visible in the motion likelihood map, but are incorrectly detected as moving by FisheyMODNet. In row (II) the walking pedestrian is correctly detected by both methods, but the dog is visible only in the motion likelihood map. It is likely that these false positives and false negatives in FisheyeMODNet are due to over-emphasis of appearance cues. In the last row (III), FisheyeMODNet correctly detects the reversing vehicle on the left. This vehicle is moving with the same direction and speed as the ego-vehicle, which is also reversing, and since it falls into the special case between the positive depth and positive height constraints it is not detected geometrically.

\subsection{Objective results}

\begin{figure}[!tbp]
  \centering
  \includegraphics[width=\columnwidth]{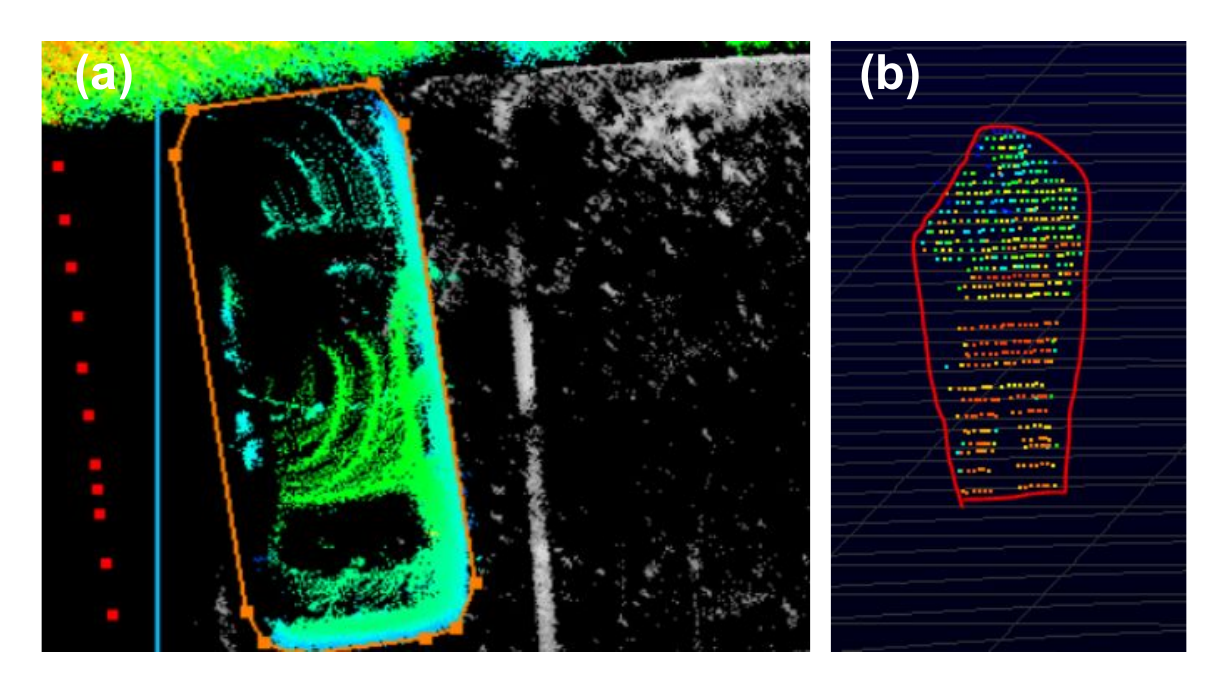}
  \vspace{-1cm}
  \caption{Ground truth generation. (a) Moving vehicle annotated in the lidar point cloud (annotation is orange polygon), and (b) point cloud corresponding to moving pedestrian projected to the fisheye image space with bounding reference polygon (red).}
  \label{fig:gt}
  \vspace{-0.5cm}
\end{figure}

In order to give objective results, a ground truth is required. As well as the four cameras, our test vehicle is equipped with a Velodyne HDL-64E lidar system. The lidar system is fully calibrated, so the geometric relationship between the lidar and the cameras is known. Moving objects are manually annotated in the lidar point cloud (\ref{fig:gt}(a)), thus identifying the point cloud features that can be associated with a dynamic object. As the camera and ground truth system are calibrated, these can then be projected to the fisheye image space (Figure \ref{fig:gt}(b)), and a bounding reference polygon is generated.

\begin{figure}[!tbp]
  \centering
  \includegraphics[width=\columnwidth]{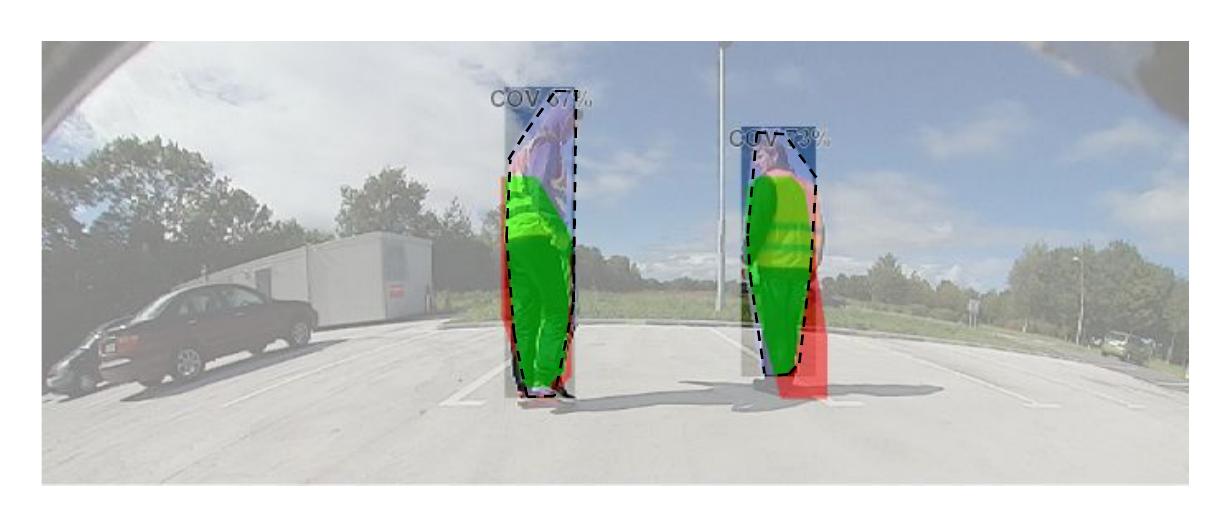}
  \vspace{-1cm}
  \caption{Detection rates explained. The ground truth is in dashed lines, the green areas show the true positive coverage of the detection, and the red shows the false positive coverage areas}
  \label{fig:detectionrates}
  \vspace{-0.5cm}
\end{figure}

Figure \ref{fig:detectionrates} shows how the true positive coverage rate (TPR) is calculated. The TPR is the ratio of the ground truth polygon area that is intersected by a detection compared to the total ground truth area, which can be expressed with the formula

\begin{equation}
    \label{eq:tpr}
    TPR = \frac{TP}{TP + FN}
\end{equation}

where TP stands for true positive and FN for false negative coverage of the total of the object area in the image. The false positive coverage is the ratio of the area of the detection that is outside the ground truth compared to the total area of the image (FP). Another metric commonly used is the intersection over union (IoU), which is defined as

\begin{equation}
    \label{eq:iou}
    IoU = \frac{TP}{TP + FP + FN}
\end{equation}

Results of the detected objects are presented in Table \ref{table_2}. 5000 frames from front cameras were used to build this data, with scenes including parking and low-speed highway. The frame rate of the camera was 15 frames per second, with speed of the vehicle ranging from 0 kph to 50kph. As we are dealing with low speed scenarios, and due to the limited effective depth of fisheye cameras, we only consider objects within 8m. Detection rate is calculated as the percentage of frames an object is detected with any coverage, while the TPR is the average coverage achieved by the detections as defined in equation \ref{eq:tpr}. The case of the static ego-vehicle (degenerate case) is in a separate row.

In Table \ref{table_2} it can be seen that preceding and approaching objects show lower percentages of coverage than the other categories and always below $50\%$, as they can be detected only with constraints applied below the horizon line. Preceding objects perform worse than approaching ones because they include vehicles positioned in front of the ego-vehicle in the same lane, which makes them centered on the epipole (or focus of expansion) and so harder to be detected. Overtaking objects have the highest detection rate as they are exclusively constituted by vehicle with mostly significantly higher speed than the ego-vehicle which makes them easier to be modelled by the dense optical flow algorithm against a slow moving background. Crossing objects have a lower detection rate and TPR as they also include crossing pedestrians, which are harder to detect in full because of their smaller speed and size in addition with non-coherent movement of the different body parts.

\begin{figure*}[t]
  \centering
  \includegraphics[width=0.9\textwidth]{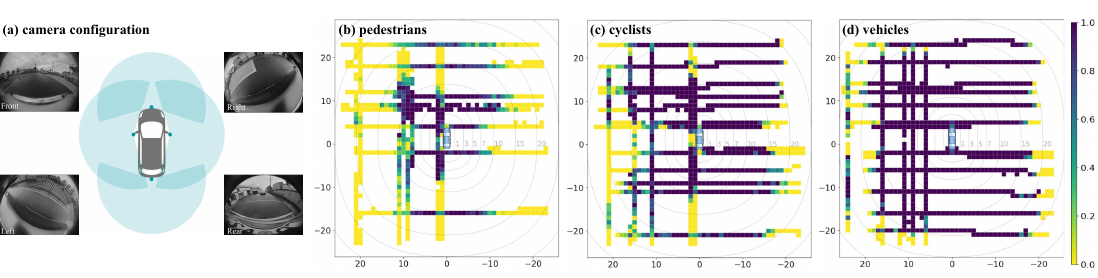}
  \vspace{-0.5cm}
  \caption{Detection rate maps for different object distances. All four cameras on the vehicle were used. The ranges are in metres. The maps are not dense, as a limited number of scenes were staged with longitudinal and lateral motion relative to the host vehicle.}
  \vspace{-0.35cm}
  \label{fig:detRanges}
\end{figure*}

An important factor that plays an important role in the detection performance is the speed. In general, a higher absolute speed of the object will provide a better detection. This can be seen in Figure \ref{fig:detRanges}, where it can be assumed that in general a vehicle moves faster than a cyclist, which in turn moves faster in general than a pedestrian. It is to be noted, however, that an upper limit to the detectable speed is given by the algorithm used for the optical flow. If an object moves too much in the camera view between frames, it can exceed the maximum search radius of the optical flow algorithm or change its own appearance in the image, causing the optical flow to fail to provide a valid result. Given the fact that the absolute and relative speed, frame rate, relative direction and distance all contribute to the final performance, it is difficult to provide a definitive analysis of the performance of the method based on the speed.

\begin{table}[!b]
\vspace{-0.4cm}
\caption{Results by category of objects} 
\vspace{-0.4cm}
\label{table_2}
\begin{center}
\renewcommand{\arraystretch}{1.5}
\begin{tabular}{c | c c c c }
\hline
\textbf{Type} 
& \textbf{\# Frames} 
& \textbf{Detection rate} & \textbf{TPR} & \textbf{IoU}\\
\hline
Crossing (epipolar)     & 3848 & 72\% & 64\% & 55\% \\ \hline
Overtaking (pos. depth)  & 2757 & 98\% & 81\% & 70\% \\ \hline
Preceding (pos. height)  & 789  & 48\% & 30\% & 19\% \\ \hline
Approaching (anti par.) & 224  & 89\% & 42\% & 30\%  \\ \hline
Static (degenerate)     & 475  & 95\% & 78\% & 69\% \\ \hline
\textbf{False Positives} 
& 5000 & 13\% &  \\
\hline
\end{tabular}
\end{center}
\vspace{-0.2cm}
\end{table}

The use of lidar as a ground truth also allows us to examine the range of detections of different object classes. A small set of staged scenes were captured, in which videos of different object types (pedestrian, cyclist, vehicle) with lateral and longitudinal relative motion were captured and processed by the proposed algorithm. The results are presented in Figure \ref{fig:detRanges}. All four cameras of a surround view system (Figure \ref{fig:detRanges}(a)) were employed. As the set of scenes is relatively small, the maps are not dense. Pedestrians naturally have a smaller good detection area, as they typically have a lower velocity than the other two object types, and thus induce smaller flow vectors for a given distance in the video.

\section{Conclusion}

We have described four geometric constraints for motion segmentation in fisheye imagery by considering the spherical geometry of the constraints. Specifically, the constraints that have been described are the epipolar, positive depth, positive height and three-view constraint. However, as we have demonstrated, there still exists a type of obstacle motion that remains undetected by these constraints - that is, specular motion compared to the host vehicle. To address this class of moving obstacle, we added the anti-parallel approach. This has a drawback of systematic false positives on close, high obstacles, but may be acceptable depending on the application. A weighted mean of the responses of the different constraints provides the final motion likelihood estimate. While the three-view constraint is discussed for the sake of completeness, it does not form part of the results as the three-view constraint requires observation of features from three camera positions, but such a correspondence is not available with dense optical flow.


The results presented, based on dense optical flow, show that the geometric approaches described are effective at detecting arbitrary moving objects. In particular, in comparison with FisheyeMODNet \cite{yahiaoui2019}, it can be seen that there are classes of objects that are detected by the geometric approaches that are undetected by the neural network approach. However, the converse is also true. Additionally, recent work in the combination of epipolar principles with data driven approaches has shown promise \cite{shen2019}, though it has not been applied to dynamic object detection. It is therefore the belief of the authors that the integration of the geometric constraints described in this paper into a neural network framework would yield optimal results.

\section*{Acknowledgment}
Many thanks to Julian Feijoo (Centro Tecnol\'{o}xico de Automoci\`{o}n de Galicia), J\'{a}n Uli\v{c}n\'{y} (Valeo Bietigheim), Fiachra Collins (Valeo Ireland) and William O'Grady (Valeo Ireland), for providing some of the results of our method. Thanks also to Senthil Yogamani (Valeo Ireland), Martin Glavin (National University of Ireland, Galway) and John McDonald (Maynooth University) for providing detailed reviews ahead of submission.

\ifCLASSOPTIONcaptionsoff
  \newpage
\fi

\bibliographystyle{IEEEtran}{}
\bibliography{references}{}

%

\begin{IEEEbiography}[{\includegraphics[width=1in,height=1.25in,clip,keepaspectratio]{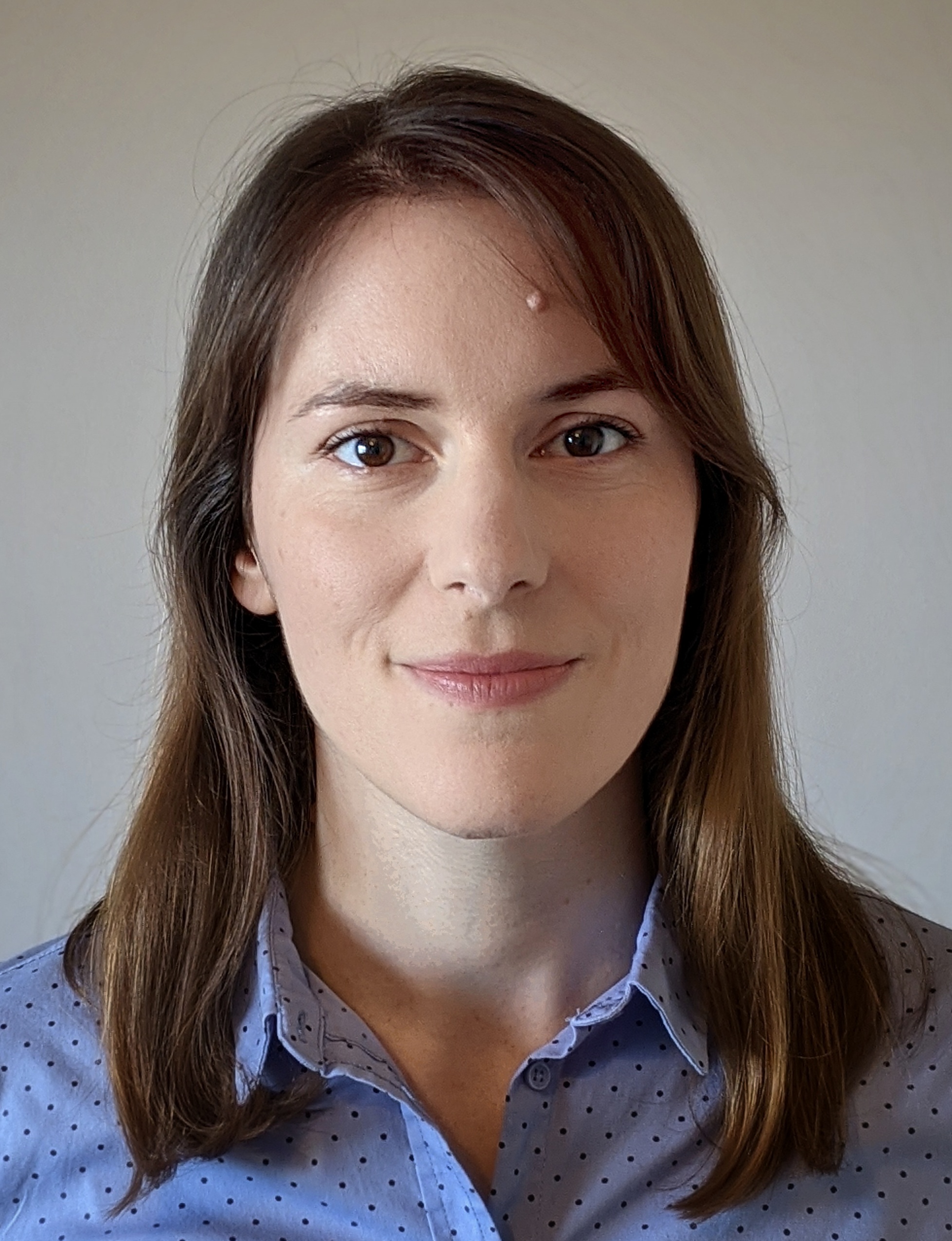}}]{Letizia Mariotti} completed her degree in Physics in 2010 and her Master of Science in Astrophysics in 2013 at the University of Trieste (Italy). She subsequently achieved her PhD in Physics from the National University of Ireland, Galway, in 2017. Letizia has been working as a computer vision research engineer in Valeo Vision Systems since 2017.%
\end{IEEEbiography}

\begin{IEEEbiography}[{\includegraphics[width=1in,height=1.25in,clip,keepaspectratio]{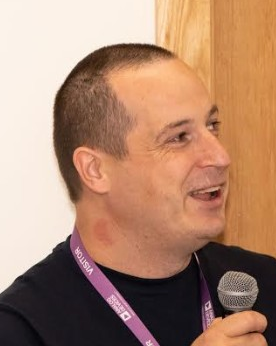}}]{Ciar\'{a}n (Hughes) Eising} completed his degree in Electronic and Computer Engineering in the National University of Ireland in 2003, and obtained his PhD from the same institute in 2010. From 2009 to 2020, Ciar\'{a}n has worked as a computer vision team lead and architect in Valeo Vision Systems, where he also held the title of Senior Expert. In 2016, he was awarded the position of Adjunct Lecturer in the National University of Ireland, Galway. In 2020, Ciar\'{a}n joined the University of Limerick.
\end{IEEEbiography}





\end{document}